\documentclass[conference]{IEEEtran}
\usepackage[utf8]{inputenc}
\usepackage{cite}
\usepackage{amsmath,amssymb,amsfonts}
\usepackage{algorithmic}
\usepackage{graphicx}
\usepackage{textcomp}
\usepackage{xcolor}

\usepackage{url}
\usepackage{booktabs}
\usepackage{algorithm}
\def\BibTeX{{\rm B\kern-.05em{\sc i\kern-.025em b}\kern-.08em
    T\kern-.1667em\lower.7ex\hbox{E}\kern-.125emX}}
\begin{document}

\title{nCMD: Benign-Anchored Feature Selection for Imbalanced Network Intrusion Detection\\
}

\author{
\IEEEauthorblockN{1\textsuperscript{st} Abu Fuad Ahmad}
\IEEEauthorblockA{\textit{Department of Computer Science} \\
\textit{New Mexico State University}\\
Las Cruces, New Mexico, USA\\
fuad@nmsu.edu}
\and
\IEEEauthorblockN{2\textsuperscript{nd} Istiaque Ahmed}
\IEEEauthorblockA{\textit{Graduate School of Informatics} \\
\textit{Osaka Metropolitan University}\\
Osaka, Japan\\
istiaque@omu.ac.jp} %sw23837u@st.omu.ac.jp}
}

\maketitle

\begin{abstract}
Feature selection is critical for network intrusion detection systems (NIDS) operating under high-dimensional, highly imbalanced traffic, as found in operational and defense networks. Traditional filter methods rank features using global statistics computed symmetrically across classes, and thus fail to capture the asymmetry of intrusion detection, where attacks are best characterized as deviations from dominant benign traffic. We propose benign-anchored Classwise Mean Deviation (nCMD), a lightweight and interpretable method that scores feature relevance by the deviation of attack-class distributions from the benign-class mean rather than a globally biased reference, aligning selection with the operational semantics of NIDS at no additional computational cost. Across four benchmark datasets—CICIDS2017, CICDDoS2019, NSL-KDD, and UNSW-NB15—multiple feature budgets, and three downstream classifiers, nCMD matches or exceeds classical filter baselines in macro-averaged F1-score, attaining the best result on three of the four datasets and under every classifier, with its advantage most pronounced under tight feature budgets and severe class imbalance. These results support benign-anchored ranking as a scalable, interpretable preprocessing component for resource-constrained NIDS.
\end{abstract}


%#############################################################
\section{Introduction}
\label{sec:introduction}
The rapid growth of network traffic generated by modern digital infrastructures has resulted in massive, high-dimensional datasets that are characteristic of big data environments. These datasets are widely used in machine learning-based network security applications; however, their large scale, feature redundancy, and frequent occurrence of missing values pose significant challenges for efficient data analysis and model development\cite{ahmad2024survayMVI, ahmad2024impactMVI}. Network intrusion detection systems (NIDS) operate in a fundamentally asymmetric regime: traffic is high-dimensional and overwhelmingly benign, while attacks are sparse and diverse, ranging from high-volume denial-of-service floods to low-rate reconnaissance \cite{Sharafaldin2018CICIDS2017,Sharafaldin2019CICDDoS2019,SommerPaxson2010}. This extreme class imbalance is intrinsic to operational networks, not an artifact of datasets, and is a long-standing challenge for learning-based detection \cite{He2009Imbalanced,BuczakGuven2016Survey}. The challenge is amplified in defense and tactical environments, where detection must run on resource-constrained edge nodes, meet tight latency budgets, and remain interpretable enough for operators to trust. In such settings benign baselines are abundant while curated attack samples are scarce, making lightweight, interpretable preprocessing that preserves attack-relevant signal especially valuable.

Among feature selection strategies, \emph{filter-based} methods are widely adopted in security pipelines for their efficiency, model independence, and compatibility with real-time constraints \cite{li2018feature,bolon2018scalability,Ring2019Survey}. Classical filter methods, including Variance Thresholding, Mutual Information, Pearson correlation, and Fisher-based criteria, rank features using statistics aggregated across all classes \cite{vergara2014mi,gu2012fisher}. As a result, each class contributes equally to the feature ranking process regardless of its prevalence or operational importance. In intrusion detection this assumption is misaligned with the problem: benign traffic defines the operational baseline, and attacks are best characterized as deviations from it \cite{SommerPaxson2010,BuczakGuven2016Survey}. Because global statistics are dominated by the benign majority, symmetric relevance scores can suppress signals from minority attack classes, systematically under-ranking features that capture rare but critical threats \cite{He2009Imbalanced}. Redundancy-aware extensions such as mRMR partially address feature dependency but add overhead and retain symmetric relevance definitions \cite{hall1999correlation,Peng2005mRMR}, leaving the semantic misalignment unresolved.

Recent work introduced \emph{Classwise Mean Deviation} (CMD), showing that simple mean-shift statistics are an effective relevance signal for intrusion detection while retaining linear complexity and interpretability \cite{AhmadICMLA2025}. CMD ranks features by aggregating class-wise deviations from a \emph{global} reference mean. However, under heavy imbalance this global mean is itself biased toward the benign class, diluting the contribution of minority attack classes—motivating a principled refinement.

We propose \emph{benign-anchored Classwise Mean Deviation} (nCMD), which replaces the global reference with the benign-class mean. Anchoring feature relevance to normal traffic reflects the core objective of intrusion detection: distinguishing malicious activity from benign behavior. This perspective also increases the contribution of attack-specific patterns to the feature ranking process. Crucially, this refinement comes at \emph{no additional cost}: nCMD preserves the linear-time efficiency, interpretability, and modularity of CMD, making it well suited to large-scale, real-time deployment on constrained platforms.

We evaluate nCMD against four classical filter baselines and the original CMD across multiple feature budgets, four benchmark datasets—CICIDS2017 \cite{Sharafaldin2018CICIDS2017}, CICDDoS2019 \cite{Sharafaldin2019CICDDoS2019}, NSL-KDD \cite{nslkdd2009}, and UNSW-NB15 \cite{moustafa2015unswnb15}—and three downstream classifiers. nCMD matches or exceeds these baselines in macro-averaged F1-score, attaining the best result on three of the four datasets and under every classifier, with its advantage most pronounced under tight feature budgets and severe imbalance—the regime most relevant to constrained operational deployment.

\textbf{Contributions.}
\begin{itemize}
    \item We propose \emph{benign-anchored Classwise Mean Deviation} (nCMD), aligning feature relevance with the asymmetric semantics of intrusion detection by measuring deviations from benign traffic, at no additional cost over symmetric mean-deviation scoring.

    \item We evaluate nCMD across four benchmark datasets, multiple feature budgets, and three downstream classifiers against classical filter baselines and CMD, showing it attains the best macro-F1 on three of four datasets and under every classifier.

    \item We provide a lightweight, interpretable, and efficient method suited to resource-constrained NIDS deployment, and release the implementation for reproducibility.\footnote{Code: \url{https://github.com/FuadAhmad/nCMD}}
\end{itemize}

\section{Related Work}
Big data applications frequently involve datasets containing thousands of features, substantial redundancy, and incomplete observations, posing significant challenges for machine learning models \cite{ahmad2021review,sayeed2024smartic}.
Feature selection has been extensively studied as a key preprocessing step for high-dimensional data analysis. Filter-based approaches, including mutual information, correlation, and Fisher-style methods, are widely used due to their computational efficiency and model independence \cite{li2018feature,vergara2014mi,gu2012fisher}. While effective in general machine learning tasks, these methods rely on global statistical measures that assume balanced class distributions.

In the context of network intrusion detection, feature selection plays a critical role in improving detection performance and scalability \cite{BuczakGuven2016Survey,Ring2019Survey,Westphal2024NIDSFS}. However, the highly imbalanced nature of network traffic presents significant challenges, as minority attack classes are often underrepresented \cite{He2009Imbalanced}. Traditional symmetric feature selection criteria may therefore fail to capture subtle deviations associated with rare but critical attack patterns \cite{SommerPaxson2010}.
Recent work introduced Classwise Mean Deviation (CMD), which leverages mean-shift statistics for feature ranking in intrusion detection \cite{AhmadICMLA2025}. While effective and computationally efficient, CMD relies on a global reference mean and thus remains influenced by class imbalance.

In contrast, this work proposes a benign-anchored formulation that explicitly aligns feature relevance with deviations from normal traffic behavior, addressing the semantic asymmetry inherent in intrusion detection tasks.

%#############################################################
\section{Methodology}
\label{sec:methodology}
The proposed nCMD method extends our earlier Classwise Mean Deviation (CMD) framework \cite{AhmadICMLA2025} by using benign traffic as the reference class during feature evaluation. Unlike CMD, which treats all classes symmetrically, nCMD measures the separation of each attack class from normal traffic. This modification is motivated by the intrusion-detection task, where identifying deviations from benign behavior is the primary objective. By evaluating classwise deviations relative to normal traffic, nCMD places greater emphasis on features that distinguish attack and benign activity, particularly in imbalanced datasets.

\textbf{Problem Setting.}
Let $D = \{(x_i, y_i)\}_{i=1}^{N}$ denote a labeled network traffic dataset, where each sample $x_i \in \mathbb{R}^{d}$ is a $d$-dimensional feature vector and $y_i \in \{0,1,\ldots,C\}$ is the corresponding class label. The label $y=0$ denotes benign (normal) traffic, while labels $y \in \{1,\ldots,C\}$ correspond to different attack classes. This formulation reflects the operational reality of NIDS, where benign traffic dominates and attacks are defined as deviations from normal behavior.

\textbf{Low-Variance Feature Filtering.}
Features with near-zero empirical variance provide limited discriminative information and are especially sensitive to sampling noise.
We therefore remove features whose variance falls below a threshold $\epsilon>0$:
\begin{equation}
\mathrm{Var}(\mathbf{x}_j) < \epsilon.
\end{equation}
This step reduces estimator variance and computational cost, and serves as a noise suppression stage prior to discriminative scoring.
In all experiments, $\epsilon$ is fixed across methods to ensure fair comparison.

\textbf{Benign-Anchored Classwise Mean Deviation (nCMD).}
The core idea of the proposed method is to quantify feature relevance by measuring the deviation of attack-class distributions relative to benign traffic along each feature dimension. To this end, we first compute class-wise mean vectors. Let $\mu_c \in \mathbb{R}^{d}$ denote the mean vector of class $c$, computed over all samples with label $y=c$. In particular, $\mu_0$ denotes the benign-class mean.

For each feature dimension $j \in \{1,\ldots,d\}$, we compute the normal-anchored relevance score as:
\begin{equation}
D^{(\mathrm{nCMD})}_j
\;=\;
\sum_{c=1}^{C} \left| \mu_{c,j} - \mu_{0,j} \right|.
\label{eq:ncmd}
\end{equation}
This score measures the accumulated absolute deviation of attack-class means from the benign mean along feature $j$. Features with larger $D^{(\mathrm{nCMD})}_j$ values exhibit stronger and more consistent deviations from normal behavior across attack classes and are therefore considered more informative for intrusion detection. Importantly, the resulting nCMD scores remain interpretable as class-aggregated mean-shift magnitudes relative to benign behavior.

\textbf{Feature Ranking and Selection.}
Features are ranked in descending order according to their nScores. For a given feature budget $k$, the top-$k$ ranked features are selected and used to train downstream classifiers. The score remains directly interpretable as a \emph{mean-shift magnitude}. As a univariate filter, the proposed method does not depend on any specific learning algorithm and can be applied as a modular preprocessing step in diverse NIDS pipelines. 

\textbf{Algorithm.}
Algorithm~\ref{alg:ncmd} summarizes the proposed nCMD feature selection procedure. Unlike wrapper or embedded methods, the approach does not require iterative model training and operates solely on class-wise summary statistics. The computational complexity is linear in the number of samples and features, i.e., $\mathcal{O}(Nd)$, making it well-suited for high-dimensional network traffic datasets.

\begin{algorithm}[t]
\caption{Feature Selection via Benign-Anchored Classwise Mean Deviation (nCMD)}
\label{alg:ncmd}
\begin{algorithmic}[1]
\REQUIRE Feature matrix $X \in \mathbb{R}^{N \times d}$, label vector $y$, number of features $k$
\ENSURE Selected feature subset $\mathcal{S}_k$

\STATE Remove features with variance below threshold $\epsilon$
\STATE Impute missing values using feature-wise means
\STATE Normalize features using Min--Max scaling

\FOR{$c = 0$ \textbf{to} $C$}
    \STATE Compute class-wise mean vector $\mu_c \leftarrow \mathrm{mean}(X_{y=c})$
\ENDFOR

\FOR{each feature index $j = 1$ \textbf{to} $d$}
    \STATE $D_j \leftarrow 0$
    \FOR{$c = 1$ \textbf{to} $C$}
        \STATE $D_j \leftarrow D_j + |\mu_{c,j} - \mu_{0,j}|$
    \ENDFOR
\ENDFOR

\STATE Rank features in descending order of $D_j$
\STATE Select top-$k$ features $\mathcal{S}_k$
\STATE \textbf{return} $\mathcal{S}_k$

\end{algorithmic}
\end{algorithm}

\section{Experimental Setup}
\label{sec:experimental_setup}

\subsection{Datasets}
We evaluate nCMD on four widely used intrusion detection benchmarks spanning modern large-scale and legacy settings: CICIDS2017~\cite{Sharafaldin2018CICIDS2017}, CICDDoS2019~\cite{Sharafaldin2019CICDDoS2019}, NSL-KDD~\cite{nslkdd2009}, and UNSW-NB15~\cite{moustafa2015unswnb15}.

\textbf{CICIDS2017} provides realistic traffic with diverse contemporary attacks (e.g., brute force, denial-of-service) over 88 flow-based features, and is heavily imbalanced toward benign traffic. \textbf{CICDDoS2019} focuses on diverse distributed denial-of-service variants under realistic multi-protocol conditions, with benign traffic forming the overwhelming majority. \textbf{NSL-KDD}, a refined version of KDD'99 that removes its redundancy and bias, has 41 numerical and categorical features and four attack groups (DoS, Probe, R2L, U2R); despite its age it remains a standard benchmark. \textbf{UNSW-NB15} offers a more recent benchmark with 49 features, nine attack categories, and moderate class imbalance.

\subsection{Baselines}
To rigorously assess the effectiveness of nCMD, we compare it against a set of widely used and conceptually relevant filter-based feature selection methods. All baselines are chosen for their computational efficiency, statistical grounding, and prior use in intrusion detection and high-dimensional learning, ensuring a fair and meaningful comparison.

\begin{itemize}
    \item \textbf{Variance Threshold} removes features with near-zero variance, eliminating attributes that contribute little discriminative information. This method is unsupervised and computationally inexpensive, making it a common preprocessing baseline~\cite{scikit2025variance}.

    \item \textbf{Pearson Correlation} ranks features according to the absolute value of their linear correlation with the class label. Features exhibiting stronger linear association with the target variable are considered more relevant~\cite{nasir2020pearson}.

    \item \textbf{Mutual Information} quantifies the statistical dependence between each feature and the target label, capturing both linear and non-linear relationships. Higher mutual information indicates greater predictive relevance~\cite{beraha2019mutual}.

    \item \textbf{Fisher Score} evaluates feature importance using the ratio of inter-class variance to intra-class variance, favoring features that achieve strong class separation with minimal within-class dispersion~\cite{gu2012fisher}.

    \item \textbf{Class-wise Mean Deviation (CMD)} ranks features based on the sum of absolute deviations of class means from a global reference mean~\cite{AhmadICMLA2025}. CMD serves as a direct and essential baseline in this study, as nCMD is derived from CMD by replacing the global reference with a benign-class anchor.
\end{itemize}

For each method, features are ranked and the top-$k$ features are selected for evaluation, ensuring consistent comparison across different feature budgets.

\subsection{Classification and Evaluation Protocol}
Experiments use Python 3.12 with \texttt{scikit-learn}~\cite{scikit-learn} and \texttt{pandas}, with Fisher Score implemented following~\cite{gu2012fisher}. We compare nCMD against four classical filter baselines—Variance Threshold, Pearson Correlation, Mutual Information, and Fisher Score—and the original CMD, from which nCMD is derived. For every method, the top-$k$ features are selected on the training data using identical partitions to ensure fair comparison, with a uniform variance threshold of $\varepsilon = 0.0$ removing only constant features.

To verify that gains are not model-specific, we evaluate three downstream classifiers of differing capacity: a three-hidden-layer MLP ($\{64,128,64\}$), a single-hidden-layer MLP, and a Decision Tree; the MLPs use the Adam optimizer with learning rate $10^{-3}$. Unless stated otherwise, main-text results use the three-hidden-layer MLP. For datasets with canonical splits (NSL-KDD, UNSW-NB15) we use them directly, yielding single-split estimates; otherwise (CICIDS2017, CICDDoS2019) we report the mean and standard deviation over stratified 5-fold cross-validation. We report Accuracy, Macro-F1, and Weighted-F1, emphasizing Macro-F1 for its equal weighting of all classes and sensitivity to minority attacks. All experiments run on an Intel Core i9-10980XE CPU with an NVIDIA RTX A4000 GPU.

\section{Results and Discussion}
\label{sec:results}

\subsection{Performance Under Tight Feature Budgets}

We first evaluate feature selection under extremely constrained settings, where only a small number of features are available ($k=5$). Tables~\ref{tab:five_feature_comparison_2017} and~\ref{tab:five_feature_comparison_2019} report classification performance on CICIDS2017 and CICDDoS2019, respectively, using the three-hidden-layer MLP.

\begin{table}[htbp]
\centering
\caption{Classification performance under a tight feature budget ($k=5$) on the highly imbalanced CICIDS2017 dataset.}
\label{tab:five_feature_comparison_2017}
\begin{tabular}{|c|c|c|c|}
\hline
\textbf{Method} & \textbf{Accuracy} & \textbf{Macro Avg} & \textbf{Weighted Avg} \\
\hline
Variance Threshold & 0.8792 & 0.3554 & 0.8427 \\
Correlation        & 0.8846 & 0.3974 & 0.8737 \\
Mutual Info.       & \textbf{0.9388} & 0.3937 & \textbf{0.9298} \\
Fisher             & 0.8780 & 0.3119 & 0.8395 \\
CMD                & 0.9146 & 0.4563 & 0.9064 \\
Our Method         & 0.9151 & \textbf{0.4611} & 0.9077 \\
\hline
\end{tabular}
\end{table}

\begin{table}[htbp]
\caption{DDoS-focused classification performance comparison of feature selection methods using a tight feature budget ($k=5$) on the CICDDoS2019 dataset.}
\label{tab:five_feature_comparison_2019}
\begin{center}
\begin{tabular}{|c|c|c|c|}
\hline 
\textbf{Method} & \textbf{Accuracy} & \textbf{Macro Avg} & \textbf{Weighted Avg} \\
\cline{1-4}
Variance Threshold & 0.8575 & 0.5616 & 0.8540 \\
Correlation        & 0.9481 & 0.6685 & 0.9433 \\
Mutual Info.       & 0.9723 & 0.6782 & 0.9680 \\
Fisher             & 0.9716 & 0.7244 & 0.9676 \\
CMD                & 0.9701 & 0.7244 & 0.9660 \\
Our Method         & \textbf{0.9744} & \textbf{0.7278} & \textbf{0.9704} \\
\hline
\end{tabular}
\end{center}
\end{table} 
On CICIDS2017, nCMD achieves the highest Macro-F1 ($0.4611$), edging out CMD ($0.4563$) and substantially exceeding the classical filters, the strongest of which (Pearson Correlation) reaches only $0.3974$. Mutual Information attains the highest accuracy but a markedly lower Macro-F1, reflecting a bias toward the dominant benign class; nCMD instead maintains balanced performance across classes, indicating greater sensitivity to minority attack categories—the property that matters most under severe imbalance.

On CICDDoS2019, nCMD attains the best score on all three metrics, including Macro-F1 ($0.7278$), narrowly ahead of CMD and Fisher Score (both $0.7244$). The margins here are small: the leading methods cluster within roughly $0.005$ Macro-F1, which is expected given that CICDDoS2019 is dominated by a few high-volume DDoS classes that are strongly separable from benign traffic regardless of the reference anchor. Even so, anchoring relevance to benign behavior is never detrimental at this budget and yields the top result.

\subsection{Performance Across Varying Feature Budgets}

\begin{figure}[htbp]
\centering
\includegraphics[width=0.95\linewidth]{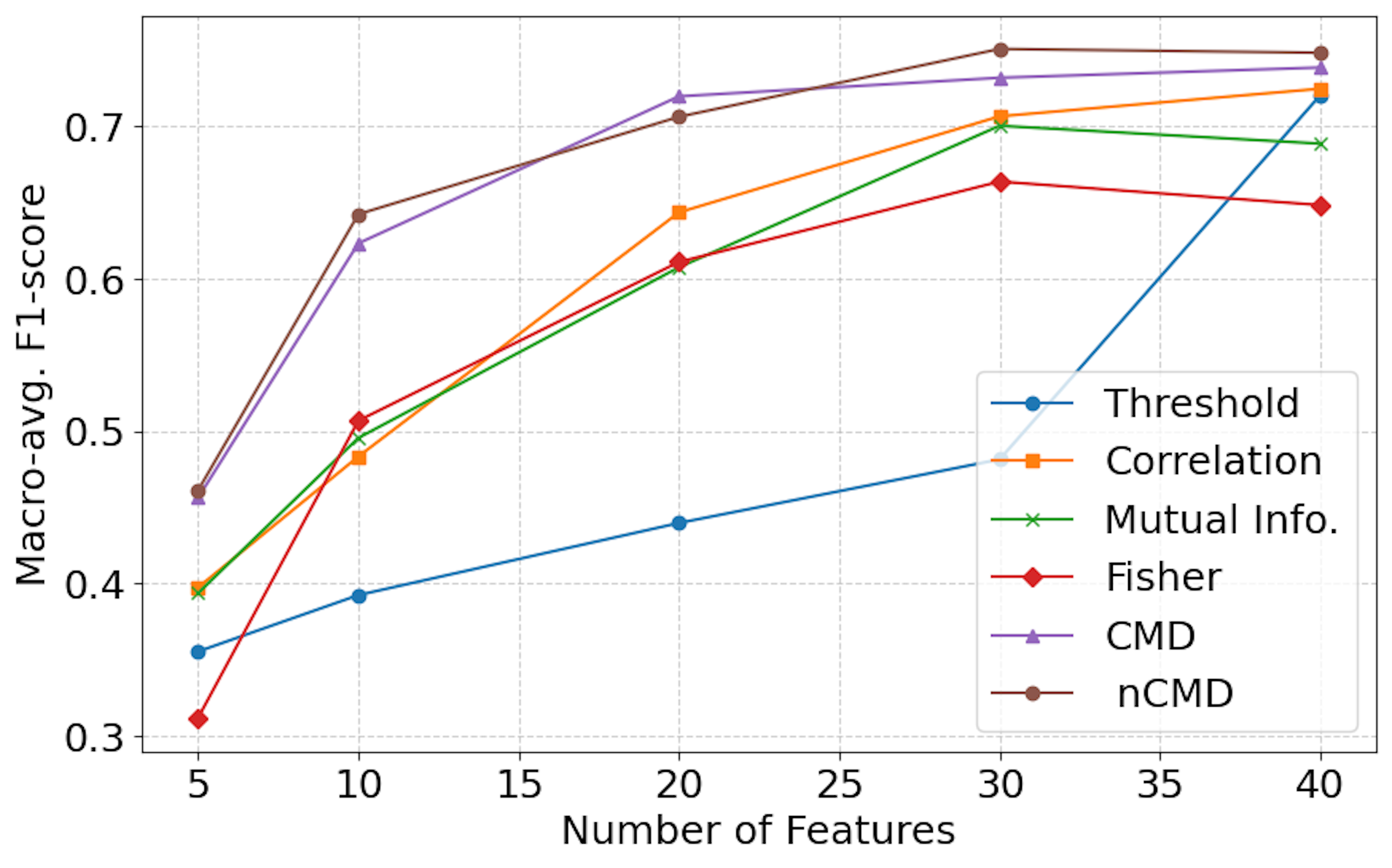}
\caption{Macro-averaged F1-score across feature subset sizes ($k=5$–40) for the proposed and baseline methods on the highly imbalanced CICIDS2017 dataset.}
\label{fig:macro_f1_cicids2017}
\end{figure}

To assess robustness under different resource constraints, we evaluate all methods across feature subset sizes from $k=5$ to $k=40$ on CICIDS2017. Table~\ref{tab:macro_f1_cicids2017} reports Macro-F1 at each budget, and Fig.~\ref{fig:macro_f1_cicids2017} visualizes the trends.

\begin{table*}[t] 
\centering
\caption{Macro-averaged F1-score comparison across different feature selection methods and varying numbers of selected features (5–40) on the highly imbalanced CICIDS2017 dataset.}
\label{tab:macro_f1_cicids2017}
\begin{tabular}{|c|c|c|c|c|c|c|}
\hline
\#Number of Features & Variance Threshold & Pearson Correlation & Mutual Information & Fisher Score & CMD & Proposed Method \\
\hline
40 & 0.7202 & 0.7245 & 0.6885 & 0.6482 & 0.7384 & \textbf{0.7481} \\
30 & 0.4814 & 0.7064 & 0.7002 & 0.6636 & 0.7317 & \textbf{0.7506} \\
20 & 0.4397 & 0.6434 & 0.6074 & 0.6109 & \textbf{0.7195} & 0.7061 \\
10 & 0.3924 & 0.4833 & 0.4955 & 0.5068 & 0.6231 & \textbf{0.6422} \\
5  & 0.3554 & 0.3974 & 0.3937 & 0.3119 & 0.4563 & \textbf{0.4611} \\
\hline
\end{tabular}
\end{table*}

nCMD attains the best Macro-F1 at four of the five budgets, and is the strongest of the two mean-deviation methods at every budget except $k=20$, where CMD leads by $0.0134$. The gap between the two is otherwise consistently in nCMD's favor (e.g., $+0.0189$ at $k=30$ and $+0.0191$ at $k=10$), indicating that benign anchoring improves feature ranking in both tightly constrained and more expressive regimes. Critically, both mean-deviation methods dominate the classical filters across the entire range: at $k=5$, nCMD reaches $0.4611$ Macro-F1 versus $0.3974$ for the best classical baseline, a margin that is operationally meaningful when feature budgets are dictated by edge-device constraints.

The classical filters also exhibit markedly less stable behavior across budgets. Variance Thresholding, in particular, degrades sharply as features are removed—from $0.7202$ at $k=40$ to $0.4397$ at $k=20$—indicating that purely unsupervised, symmetric criteria can fail to preserve discriminative features when the subset is small. By contrast, the benign-anchored ranking retains attack-relevant features more reliably under tight budgets, which is the regime of greatest practical interest for deployable NIDS.

\subsection{Cross-Dataset Generalization}
\label{sec:cross_dataset}

We evaluate the generalization of nCMD across all four benchmarks using the top-40 selected features and a three-hidden-layer MLP, with results summarized in Table~\ref{tab:cross_dataset}. nCMD attains the highest macro-F1 on three of the four datasets—CICIDS2017, NSL-KDD, and UNSW-NB15—and remains competitive on the fourth. Relative to CMD, its closest baseline and the method it directly refines, nCMD improves macro-F1 on three datasets, with the largest gain on the legacy NSL-KDD benchmark ($0.3124 \rightarrow 0.3429$, $+0.0305$) and consistent gains on CICIDS2017 ($+0.0097$) and UNSW-NB15 ($+0.0094$).

The single exception is CICDDoS2019, where nCMD ($0.7276$) trails CMD ($0.7320$) by a narrow margin. Notably, this dataset is dominated by a small number of high-volume DDoS classes whose distributions differ strongly from benign traffic regardless of the reference point; here the top filter methods (CMD, Fisher, and nCMD) cluster within $0.005$ macro-F1 of one another, indicating that the choice of anchor has limited effect when attack classes are large and clearly separable. The benefit of benign anchoring is instead most evident on datasets with greater attack-class diversity and more severe minority underrepresentation, such as NSL-KDD and UNSW-NB15, which is consistent with the motivation of the method.

Overall, these results indicate that nCMD generalizes across both modern (CICIDS-series) and legacy (NSL-KDD, UNSW-NB15) benchmarks, and that benign anchoring yields the clearest advantage precisely in the diverse, highly imbalanced regimes that characterize operational network defense.

\begin{table*}[htbp] 
\centering
\caption{Macro-averaged F1-score (mean $\pm$ std over 5 fold CV) using the top-40 selected features with a three-hidden-layer MLP across four benchmark intrusion detection datasets. NSL-KDD and UNSW-NB15 use their canonical predefined test splits, so variance is zero. Best result per dataset in \textbf{bold}.}
\label{tab:cross_dataset}
\begin{tabular}{|c | c| c| c |c|} 
\toprule
\textbf{Method} & \textbf{CICIDS2017} & \textbf{CICDDoS2019} & \textbf{NSL-KDD} & \textbf{UNSW-NB15} \\
\midrule

Variance Threshold & 0.7202 $\pm$ 0.0081 & 0.7230 $\pm$ 0.0047 & 0.2835 & 0.3436 \\
Pearson Correlation & 0.7245 $\pm$ 0.0108 & 0.7275 $\pm$ 0.0042 & 0.2881 & 0.3246 \\
Mutual Information & 0.6885 $\pm$ 0.0125 & 0.7269 $\pm$ 0.0046 & 0.3274 & 0.3670 \\
Fisher Score & 0.6482 $\pm$ 0.0121 & \textbf{0.7304} $\pm$ 0.0049 & 0.3209 & 0.3652 \\
CMD & 0.7384 $\pm$ 0.0156 & \textbf{0.7320} $\pm$ 0.0041 & 0.3124 & 0.3646 \\
\textbf{nCMD (Proposed)} & \textbf{0.7481} $\pm$ 0.0183 & 0.7276 $\pm$ 0.0050 & \textbf{0.3429} & \textbf{0.3740} \\
\bottomrule
\end{tabular}
\end{table*}

\subsection{Robustness Across Downstream Classifiers}
\label{sec:classifier_robustness}

Because feature selection is a preprocessing step, a useful selector should improve performance independently of the downstream model. To test this, we re-evaluate all methods on CICIDS2017 using three classifiers of differing capacity: a Decision Tree, a single-hidden-layer MLP, and the three-hidden-layer MLP used elsewhere in this work (Table~\ref{tab:classifier_robustness}). nCMD achieves the highest macro-F1 under every classifier, and improves over CMD in all three cases ($+0.0052$ for the Decision Tree, $+0.0097$ for the 3-layer MLP, and $+0.0146$ for the 1-layer MLP). The advantage is largest for the lowest-capacity model, suggesting that benign-anchored ranking is particularly beneficial when the downstream classifier cannot compensate for poorly chosen features—an operationally relevant setting for lightweight, edge-deployed detectors.

\begin{table*}[htbp]
\centering
\caption{Macro-averaged F1-score (mean $\pm$ std) on CICIDS2017 with top-40 selected features across three downstream classifiers. nCMD ranks first under every classifier, indicating that its advantage is not tied to a specific model.}
\label{tab:classifier_robustness}
\begin{tabular}{|c|c|c|c|}
\hline
\textbf{Method} & \textbf{Decision Tree} & \textbf{MLP (3 hidden layer)} & \textbf{MLP (1 hidden layer)} \\
\hline
Variance Threshold & 0.8898 ± 0.0177 & 0.7202 ± 0.0081 & 0.5884 ± 0.0293 \\
Correlation        & 0.8917 ± 0.0208 & 0.7245 ± 0.0108 & 0.6898 ± 0.0178 \\
Mutual Info.       & 0.8891 ± 0.0131 & 0.6885 ± 0.0125 & 0.6290 ± 0.0466 \\
Fisher             & 0.8531 ± 0.0216 & 0.6482 ± 0.0121 & 0.6607 ± 0.0370 \\
CMD                & 0.8936 ± 0.0087 & 0.7384 ± 0.0156 & 0.6814 ± 0.0304 \\
nCMD (Proposed)    & \textbf{0.8988 ± 0.0210} & \textbf{0.7481 ± 0.0183} & \textbf{0.6960 ± 0.0196} \\
\hline
\end{tabular}
\end{table*}

\subsection{Discussion}
The empirical results support the central hypothesis of this work: feature relevance in intrusion detection is inherently asymmetric and is well measured relative to benign traffic behavior. By anchoring classwise mean deviation to the benign class, nCMD provides a semantically aligned scoring mechanism that improves the visibility of minority attack classes, attaining the best macro-F1 on three of the four benchmarks and on all three downstream classifiers.

The pattern of results also clarifies \emph{when} benign anchoring helps. Its advantage is largest on datasets with diverse attack classes and severe minority underrepresentation, such as NSL-KDD and UNSW-NB15, and under tight feature budgets where the choice of relevance criterion is most consequential. Conversely, on CICDDoS2019—dominated by a few high-volume DDoS classes that are strongly separable from benign traffic—nCMD, CMD, and Fisher Score perform near-identically, indicating that the reference anchor matters little once attack classes are large and clearly distinguishable. This is consistent with the formulation: for highly imbalanced data the global mean already lies close to the benign mean, so the two scores diverge most when attack classes are numerous and individually small.

Although nCMD changes the feature-scoring strategy, it retains the low computational overhead and interpretability that make filter-based methods attractive for large-scale intrusion-detection applications. The improvement is most pronounced for the lowest-capacity classifier, suggesting that benign-anchored ranking is especially valuable when the downstream model cannot compensate for poorly chosen features, as in lightweight, edge-deployed detectors. Overall, these findings indicate that a simple, semantically grounded modification to the feature relevance criterion can improve intrusion detection performance under class imbalance without increasing model complexity.

\section{Limitations and Scope}
\label{sec:limitations}
While nCMD performs strongly across multiple intrusion detection benchmarks, several limitations define its scope.

First, nCMD is a \emph{univariate, mean-based} method, sensitive only to first-order shifts. Attacks that deviate from benign traffic through higher-order characteristics—variance, temporal dependencies, or feature interactions—may be missed, in which case wrapper or embedded methods, or more expressive classifiers, may offer complementary benefits.

Second, nCMD assumes a well-defined benign class. This holds in most supervised settings but is less applicable to unsupervised or open-world scenarios where benign labels are unavailable or evolving; extending benign anchoring to such settings remains future work.

Third, all attack classes contribute equally to the deviation score. This emphasizes rare attacks but may underweight dominant categories; class-dependent weighting could offer a tunable trade-off between sensitivity to rare and frequent attacks.

Finally, our evaluation relies on four benchmark datasets that may contain artifacts and biases, so results should be read within that context. Validation on additional sources and live network traffic would further strengthen generalization claims.

Despite these limitations, nCMD provides a simple, interpretable, and efficient feature ranking that improves detection under severe class imbalance while remaining scalable for practical deployment.

\section{Conclusion}
\label{sec:conclusion}
This paper presented \emph{benign-anchored Classwise Mean Deviation} (nCMD), a filter-based feature selection method that evaluates feature relevance relative to benign traffic. By incorporating the distinction between normal and attack behavior directly into the ranking process, nCMD improves sensitivity to attack-specific characteristics while retaining the simplicity and efficiency of conventional filter methods. Experimental results on CICIDS2017, CICDDoS2019, NSL-KDD, and UNSW-NB15 showed that nCMD consistently performs competitively and often achieves higher macro-averaged F1-scores, particularly in imbalanced settings and under limited feature budgets. These results indicate that incorporating benign traffic as a reference class can improve feature selection for imbalanced intrusion-detection tasks without increasing computational complexity.
%We proposed \emph{benign-anchored Classwise Mean Deviation} (nCMD), a lightweight, interpretable feature selection method that addresses the asymmetric, highly imbalanced nature of network intrusion detection by anchoring relevance to benign traffic. Across four benchmark datasets—CICIDS2017, CICDDoS2019, NSL-KDD, and UNSW-NB15—and three downstream classifiers, nCMD matches or exceeds classical filter methods, attaining the best macro-averaged F1-score on three of the four datasets and under every classifier, with its advantage most pronounced under tight feature budgets and severe imbalance. It achieves this while retaining the efficiency, interpretability, and ease of integration of filter-based methods at no additional computational cost, making it well suited to scalable, resource-constrained NIDS deployments such as those in defense and tactical networks.

%Future work will extend the benign-anchored formulation to streaming and open-world settings, incorporate higher-order statistics to capture more complex attack behaviors, and evaluate nCMD in operational and adversarial environments.

%##################################################################

\bibliographystyle{IEEEtran}% \bibliographystyle{plain}
\bibliography{bibliography_verified}
%\vspace{12pt}

\end{document}